\documentclass[sigplan,twocolumn]{acmart}
\renewcommand\footnotetextcopyrightpermission[1]{}
\setcopyright{none}

\usepackage{amsmath}
\usepackage{array}
\usepackage{booktabs}
\usepackage{enumitem}
\usepackage{multirow}
\usepackage{tabularx}
\usepackage{xspace}
\usepackage{algorithm}
\usepackage{algpseudocode}
\usepackage{tikz}
\usetikzlibrary{decorations.pathreplacing,arrows.meta,patterns}

\newcommand{\xmark}{$\times$}

\newcommand{\SystemName}{\textsc{ARC}\xspace}
\newcommand{\CSB}{\textsc{CSB}\xspace}
\newcommand{\DGC}{\textsc{DGC}\xspace}
\newcommand{\AAP}{\textsc{AAP}\xspace}

\newcommand{\cparagraph}[1]{\noindent\textbf{#1.}}

\newcolumntype{Y}{>{\raggedright\arraybackslash}X}

\begin{document}
\title{\SystemName: Adaptive Rollout Control for Efficient Synchronous LLM Reinforcement Learning}

\author{Yihui Zhang\textsuperscript{1}, 
Tianyu Wo\textsuperscript{1}, 
Jinghao Wang\textsuperscript{1},
Xiao Zhou\textsuperscript{1},
Huitian Xue\textsuperscript{1},
Xiaoyang Sun\textsuperscript{2\dag},
Zeying Wang\textsuperscript{1},
Chunming Hu\textsuperscript{1},
Renyu Yang\textsuperscript{1\dag}
\thanks{Corresponding Authors: Xiaoyang Sun (x.sun4@leeds.ac.uk), Renyu Yang (renyuyang@buaa.edu.cn)}
}

\affiliation{%
$^1$Beihang University \quad \quad $^4$University of Leeds
\country{}
}

\renewcommand{\authors}{Yihui Zhang, Tianyu Wo, Jinghao Wang, Xiao Zhou, Huitian Xue, Xiaoyang Sun, Zeying Wang, Chunming Hu, Renyu Yang}

\renewcommand{\shortauthors}{Y. Zhang, Tianyu Wo,  et. al}

\begin{abstract}
Synchronous reinforcement learning (RL) for large language models proceeds through a repeated rollout--barrier--update loop. Each update must wait for a fixed number of complete, group-relative training samples, yet individual response lengths and reward outcomes vary widely and only emerge as generation progresses. Static admission and unconditional completion therefore leave accelerators idle in some steps, and waste computation generating samples that will never contribute to the current update. We present \SystemName, a closed-loop rollout architecture that governs synchronous RL rollout at group granularity. \SystemName uses \textit{Calibrated Streaming Backfill} to judge from partial observations whether a group is still worth completing, backfilling a fresh candidate immediately when it is not; \textit{Deferred Group Completion} to park an uncertain early stop and resume it later instead of discarding it outright, bounded by a resume-cycle limit that guarantees termination; and \textit{Adaptive Admission} and \textit{Pipelining} to order admitted samples by early-measured duration and to admit further candidates only while expected supply still falls short of the target number of accepted groups. Under online evaluation, \SystemName reduces rollout time by $37.4\%$, a $1.60\times$ speedup, and improves effective GRPO throughput by $50.1\%$ over existing baselines, while directly recovering 98.9\% of the training groups that would ultimately have been kept.
\end{abstract}


\maketitle
\section{Introduction}
\label{sec:introduction}

Large Language Models Reinforcement Learning (LLM RL) frameworks, such as veRL~\cite{sheng2025hybridflow} and slime~\cite{slime_github}, improve model policies by repeatedly generating responses, evaluating rewards, and using that feedback to update the policy. These frameworks typically use two main engines: an \textit{inference} engine for rollout generation and a \textit{training} engine for policy updates. Coordinating the two creates a challenge: keeping hardware busy while ensuring that rollout data remain aligned with the policy being trained. ARC targets the synchronous regime, which bounds staleness at the cost of hardware utilization~\cite{sheng2026laminar,wu2025llamarl}.

\begin{figure}[t]
\centering
\includegraphics[width=\linewidth]{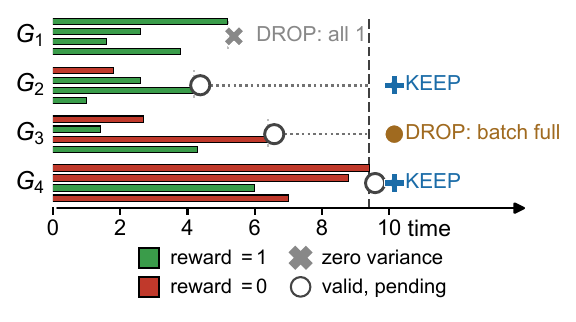}
\vspace{-1.2em}
\caption{
Sample reward accumulation. $\mathrm{Green}$ and $\mathrm{red}$ denote sample rollouts with $\mathrm{rewards}=1$ and $=0$, respectively.
Upon completion, group $G_1$ with zero-variance is rejected (\xmark), whereas the samples ($\circ$) await batch selection.
At the synchronization barrier (vertical dashed line), the valid groups that remain are retained uniformly at random until the $B$-group batch fills, here $G_2$ and $G_4$ marked \textsc{keep}, while the rest are discarded even though they are equally valid, here $G_3$ marked \textsc{drop}.
}

\vspace{-4mm}
\label{fig:motivation}
\end{figure}

Within this synchronous loop, the policy-optimization algorithm determines how rollout samples are organized and selected to provide informative learning signals for each update.
Group Relative Policy Optimization (GRPO)~\cite{shao2024deepseekmath} and its refinement, Decoupled Clip and Dynamic sAmpling Policy Optimization (DAPO)~\cite{yu2026dapo}, are now the common foundation in LLM RL: DAPO samples $S$ completions per prompt as a \emph{group}, filters out groups whose samples all receive the same reward, and oversamples extra candidate groups to still reach a fixed target of $B$ valid groups every step~\footnote{$S$ and $B$ refers to the number of responses sampled per prompt and the target number of accepted prompt groups per training batch, respectively.}. Both techniques improve training quality, but at a direct inference cost: candidates drawn beyond what a step keeps are generated and scored, yet are never used.


In addition to the total number of candidates generated, the selection policy governs which valid groups are incorporated into the training set. A first-finished-first-kept strategy decreases wall-clock latency; however, it systematically privileges groups with shorter completion times, thereby potentially underrepresenting trajectories that require longer reasoning horizons. To mitigate this completion-time–induced selection bias, the synchronous pipeline waits for all $S$ responses within each candidate group to finish before performing selection uniformly over \emph{all} valid groups. This procedure renders selection invariant to the relative completion order of valid candidates, at the cost of postponing batch finalization until the slowest rollout has terminated.

Figure~\ref{fig:motivation} illustrates this with four candidate groups whose samples complete at different times. Once all of $G_1$'s samples finish, its reward variance turns out to be zero, so it is dropped; $G_2$, $G_3$, and $G_4$ are all valid once complete, but the step targets only $B{=}2$ groups, and which two fill them emerges only at the barrier, i.e., $G_4$'s completion, since it is the slowest. The \emph{dynamic filter} randomly keeps $G_2$ and $G_4$ and drops $G_3$ (even as valid) because only $B=2$ groups are needed.

\begin{figure*}[t]
\centering
\includegraphics[width=0.93\textwidth]{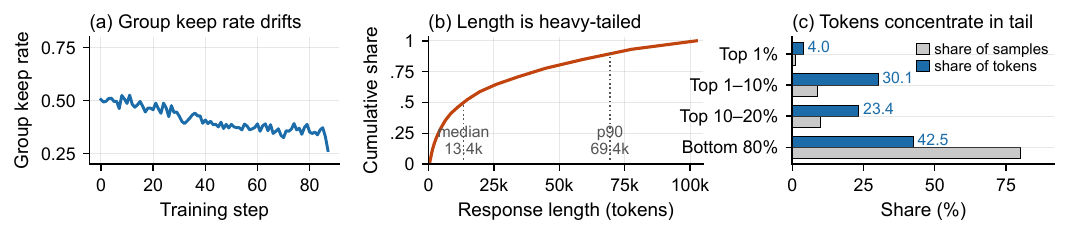}
\caption{Statistics from 88 recorded production rollout steps. (a) Per-step group keep rate. (b) CDF of sample response length at a representative step. (c) Share of samples against share of generated tokens, by length band, at the same step.}
\vspace{-0.1mm}
\label{fig:prod-obs}
\end{figure*}


Recent approaches to rollout scheduling address this barrier by reordering, regrouping, or relocating admitted work~\cite{qin2026seer,gao2025rollpacker,xu2026tailsieve,zhang2026heddle,zhou2025april}.
While these methods reduce the straggler's cost, they do not mitigate oversampling-induced inefficiency: 
groups that will eventually be discarded are nevertheless executed to completion.
A second consideration is early termination~\cite{voss2026early,zhai2026selective,wang2026darts}; however, termination is typically triggered based on a statistical estimate, suggesting that the data seen during policy optimization depends on the selected stopping rule.
In contrast, \SystemName determines, at the system level, \emph{which} candidates are worth completing and makes every stop reversible rather than permanent.
Consequently, the policy optimization is preserved: it continues to observe complete $S$-sample groups, with exactly $B$ groups per batch.


We therefore conceptualize rollout execution as a closed-loop control problem, formulated around a central question: \emph{what fraction of the rollout computation performed during a synchronous reinforcement learning (RL) update can be rendered provably effective without relaxing the synchrony constraints of the outer optimization loop or compromising the quality of the learned model?} 
We present \SystemName, an integrated framework that comprises i) a core rollout-control mechanism, ii) a reliability mechanism that supports recovery from incorrect stopping decisions, and iii) an integration layer that closes the feedback loop via adaptive admission control and scheduling.

The paper makes the following contributions:
\begin{itemize}[leftmargin=1.2em,topsep=0pt,partopsep=0pt]
    \item \textbf{Compute-aware rollout control.} Calibrated Streaming Backfill (\CSB) turns a group's partial completions into calibrated evidence, then decides by expected cost, weighing remaining generation against the cost and feasibility of replacing it before this step's barrier, and immediately admits a replacement from DAPO's oversampling pool when it stops.
    \item \textbf{Recoverable early stopping.} Deferred Group Completion (\DGC) preserves a stopped group's progress instead of discarding a potentially wrong self-learned stop. It is bounded by a resume-cycle limit that guarantees every parked group eventually reaches a real verdict, rendering \CSB's uncertainty tolerable.
    \item \textbf{Closed-loop admission and ordering.} Adaptive Admission and Pipelining (\AAP) orders already-admitted samples by pipelining and admits further candidates only while the expected supply of accepted groups still falls short of the step's batch, reusing \CSB's per-group belief as its input.
\end{itemize}

\section{Background and Motivation}
\label{sec:background}

\subsection{Synchronous Group-Based LLM RL}

A synchronous LLM RL step alternates between a rollout runtime, which generates candidate groups and their rewards under a policy, and a training runtime, which consumes the accepted groups and updates the policy. 
These two phases may be co-located or disaggregated; however, they remain separated by an operational barrier: the training phase commences only after the rollout phase has generated the corresponding trajectories.


Let $B$ represent the number of accepted groups processed at each optimization step, and let $S$ represent the number of samples collected within each group. Every sample consists of a full model generation together with its corresponding reward; in reinforcement learning terms, this is a \emph{trajectory}. Thus, each group includes $S$ trajectories.
Oversampling increases the \emph{number of candidate groups} produced in each iteration above the setting of $B$ accepted groups, while holding the group size constant at $S$. Group-relative approaches calculate advantages using reward statistics computed within each group; in general, estimating these statistics from an incomplete subset of samples does not yield equivalent advantages to those obtained from the full group. Therefore, under the fixed-size group interface assumed here, a group can be used for training only once all $S$ samples have finished and their corresponding rewards are available.


\subsection{Challenge 1: Rollout Dynamics and Heavy Tails}
\label{sec:rollout-dynamics}



Under synchronous execution, each policy update must wait until rollout produces $B$ accepted groups. Two empirical properties of the workload make satisfying this constraint more expensive: the group keep rate drifts over training, and a small number of very long responses contributes a large fraction of the total generated tokens. While continuous batching and memory-aware serving can raise hardware utilization~\cite{yu2022orca,kwon2023efficient,zheng2024sglang}, these techniques by themselves do not decide how many candidate groups to accept or which partially processed groups are worth finishing.

\cparagraph{Rollout dynamics: a drifting keep rate} How many candidates a step must draw to obtain $B$ usable groups depends on the group \textit{keep rate}. Figure~\ref{fig:prod-obs}(a) shows it over 88 production rollout steps: it drifts from $0.53$ to $0.27$, a $49\%$ relative decline with rank correlation $\rho=-0.93$ against step index. A static oversampling multiple is therefore mis-sized: large enough at one \textit{keep rate}, it overdraws once the filter becomes more permissive; tuned to the average, it leaves some steps short of $B$ and forces an extra admission round. The decline is uneven, with only about half of steps falling below their predecessor, so the series fluctuates around a moving trend, arguing for online tracking and smoothing rather than step-by-step re-estimation. Filtering compounds this: a group is usable only after all $S$ samples finish and survive the filter, so rollout should minimize not average request latency but the time when the $B$-th accepted complete group becomes available, because later generation cannot serve the current update.

\cparagraph{Heavy tails: tokens concentrate in the longest samples} Figure~\ref{fig:prod-obs}(b) shows the CDF of response lengths at a representative step. Lengths span nearly three decades, from $104$ tokens to the $100$K-token generation limit; the median is $13.4$k, the $90$th percentile is $69.4$k, and $1.3\%$ hit the limit. This skew drives token volume: Figure~\ref{fig:prod-obs}(c) groups samples by length and compares each band’s share of samples vs. tokens (they would match under uniform lengths). The top $1\%$ produces $4.0\%$ of tokens ($4.0\times$ its share), the rest of the top decile produces $30.1\%$ ($3.3\times$), so the top $10\%$ accounts for $34.1\%$; the next decile adds $23.4\%$. Conversely, the bottom $80\%$ of samples generates only $42.5\%$ of tokens ($0.53\times$). Tokens therefore concentrate in long samples, where stopping decisions matter most, rather than being evenly spread across samples th
at a scheduler could reorder.

\subsection{Challenge 2: The Ceiling of Priority Scheduling}
\label{sec:priority-ceiling}

Because variance in makespan is largely attributable to heterogeneity in response lengths, an intuitive baseline method is to prioritize requests with the longest expected completions. Seer~\cite{qin2026seer} provides a representative example: it issues an early “probe” generation for each prompt group and uses the resulting partial output as an online estimator of the group’s remaining generation length. It then approximates a longest-job-first (LJF) policy by co-batching requests predicted to be long with those predicted to be short, thereby maintaining high batch utilization.

\begin{figure}[t]
\centering
\includegraphics[width=\linewidth]{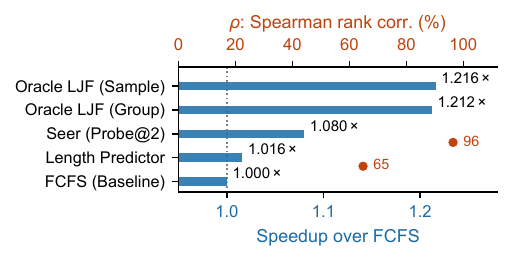}
\vspace{-1.2em}
\caption{Comparison of priority scheduling strategies under otherwise identical runtime settings.}
\vspace{-4mm}
\label{fig:priority-ceiling}
\end{figure}

Since Seer couples its ordering heuristic with additional system mechanisms, we replay a recorded production rollout trace under multiple priority signals while holding admission control and stopping criteria constant. Specifically, we compare: (i) first-come, first-served (FCFS); (ii) a static prompt-label–based length predictor; (iii) probe-based prioritization using two early samples per group (probe@2), which reimplements Seer at our GRPO group/sample granularity; and (iv) two oracle variants that reorder requests using the true generation length as an upper bound. Figure~\ref{fig:priority-ceiling} reports the speedup of each configuration relative to FCFS.

Figure~\ref{fig:priority-ceiling} isolates two points. First, reordering yields a gain, and the magnitude of that gain tracks signal quality: probe@2 reaches $\rho=0.96$ and $1.08\times$, while the static predictor, limited by prompt-level features, reaches only $\rho=0.65$ and $1.02\times$. Second, it is bounded: even an oracle with exact per-sample lengths reaches only $1.22\times$, and a group-level-length oracle, the best a real predictor could target since within-group variation is sampling randomness, already reaches $1.21\times$. This bound is structural and workload-dependent. Reordering can only fill slots left idle by length skew: a flat length profile leaves many short jobs to pack around long ones, whereas the profile in \S\ref{sec:rollout-dynamics} is tail-heavy, with the top $1\%$ of samples carrying $4.0\%$ of generated tokens and the top $10\%$ carrying $34.1\%$, leaving little to pack. Under a fixed-capacity pipeline, a dispatched batch cannot finish faster than its longest sample, and once all samples are dispatched, no work remains for reordering to rearrange; on this trace, dispatch phases run at full occupancy and remaining idle slot-time sits in the final tail. Reordering thus changes \emph{when} necessary tokens are generated, not \emph{how many} are necessary, and $1.22\times$ is the limit of what rearrangement alone can achieve. Exceeding it requires declining to finish some long tail samples, using early completions not to order work but to judge whether it is worth completing at all.

\subsection{Opportunity: Observability from Probe Sampling}
\label{sec:probe}

\S\ref{sec:priority-ceiling} measured the ceiling of one specific use of an early-completion signal, namely setting priority from a group's first two finished samples. The phenomenon behind it is broader than that one setting. A group's $S$ samples do not finish together: as decoding proceeds they complete one at a time, so after any $n$ are done the group sits in a \emph{partially observed} state, with $n$ finished, $S-n$ pending, plus whatever reward or filter outcome has emerged so far. We call this progression from $n=1$ to $n=S$ \emph{probe sampling}: not a designated probe count, but the fact that a group's own emerging results are the earliest evidence available about its eventual verdict.

This progression has two uses, and it is important not to conflate them. One is \emph{ordering}: a duration signal read from the progression, available from $n=2$ onward and refreshed as further samples complete, sets the priority of the group's remaining samples, so groups measured to run long are scheduled with higher priority. This is the mechanism whose ceiling \S\ref{sec:priority-ceiling} measured; instantiated as \emph{Probe@2 dynamic pipelining} (\S\ref{sec:aap}), it is on its own a minor contributor to makespan (\S\ref{sec:online-results}). The other, more fundamental use is \emph{informational}: at any point, the completed samples are the only evidence the runtime has about whether the rest of the group is worth finishing. Without incremental arrival of results, a group would be either fully unobserved or fully complete, with nothing in between to reason about.

\SystemName's stopping decision (\S\ref{sec:csb}) rests on the second use of probe sampling: it evaluates a group's calibrated keep probability continuously as $n$ advances, from a small warm-up count to $S-1$, following the same progression rather than relying on one fixed setting. Probe sampling is thus the entry point of the control loop. Its value there lies not in any particular sample count but in the fact that the same incremental completions that bound the reordering ceiling also make group-level early decisions possible at all. That is the key insight behind \SystemName's ability to raise \emph{effective} rollout token throughput (tokens spent on samples that end up in an accepted training group), whereas reordering already optimizes raw generated-token throughput nearly as well as is possible. The three remaining limitations are addressed in \S\ref{sec:csb} through \S\ref{sec:aap}: running every sample to completion, making an early stop irreversible, and admitting candidates only at fixed boundaries.

\section{Design of \SystemName}
\label{sec:design}

\begin{figure}[tb]
\centering
\includegraphics[width=\linewidth]{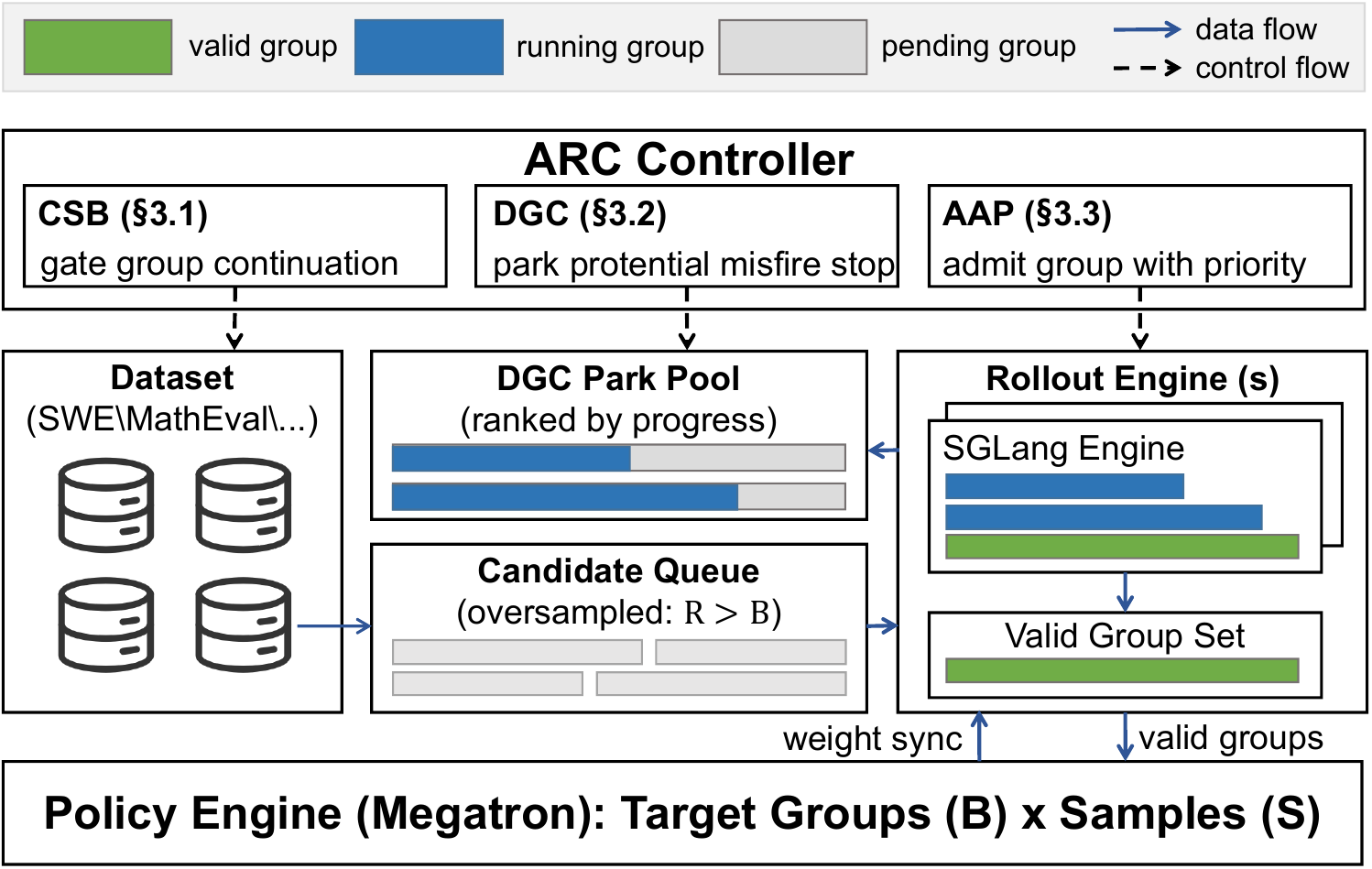}
\caption{Overview of \SystemName.}
\vspace{-4mm}
\label{fig:arc-overview}
\end{figure}

To reduce step-level rollout makespan and wasted generation as dynamics drift, \SystemName separates three control scopes (Figure~\ref{fig:arc-overview}) that operate on probe sampling’s partial observability (\S\ref{sec:probe}). \CSB decides whether an active group should continue; \DGC retains stopped work for possible reuse, mitigating \CSB’s potential stopping errors; and \AAP controls how many candidate groups enter the rollout engines and the execution order of admitted samples, including those \DGC resumes. Together, they convert probe sampling into early, recoverable group verdicts and close the loop between verdicts and admission count.

\subsection{Calibrated Streaming Backfill}
\label{sec:csb}

\begin{figure}[t]
\centering
\includegraphics[width=0.95\linewidth]{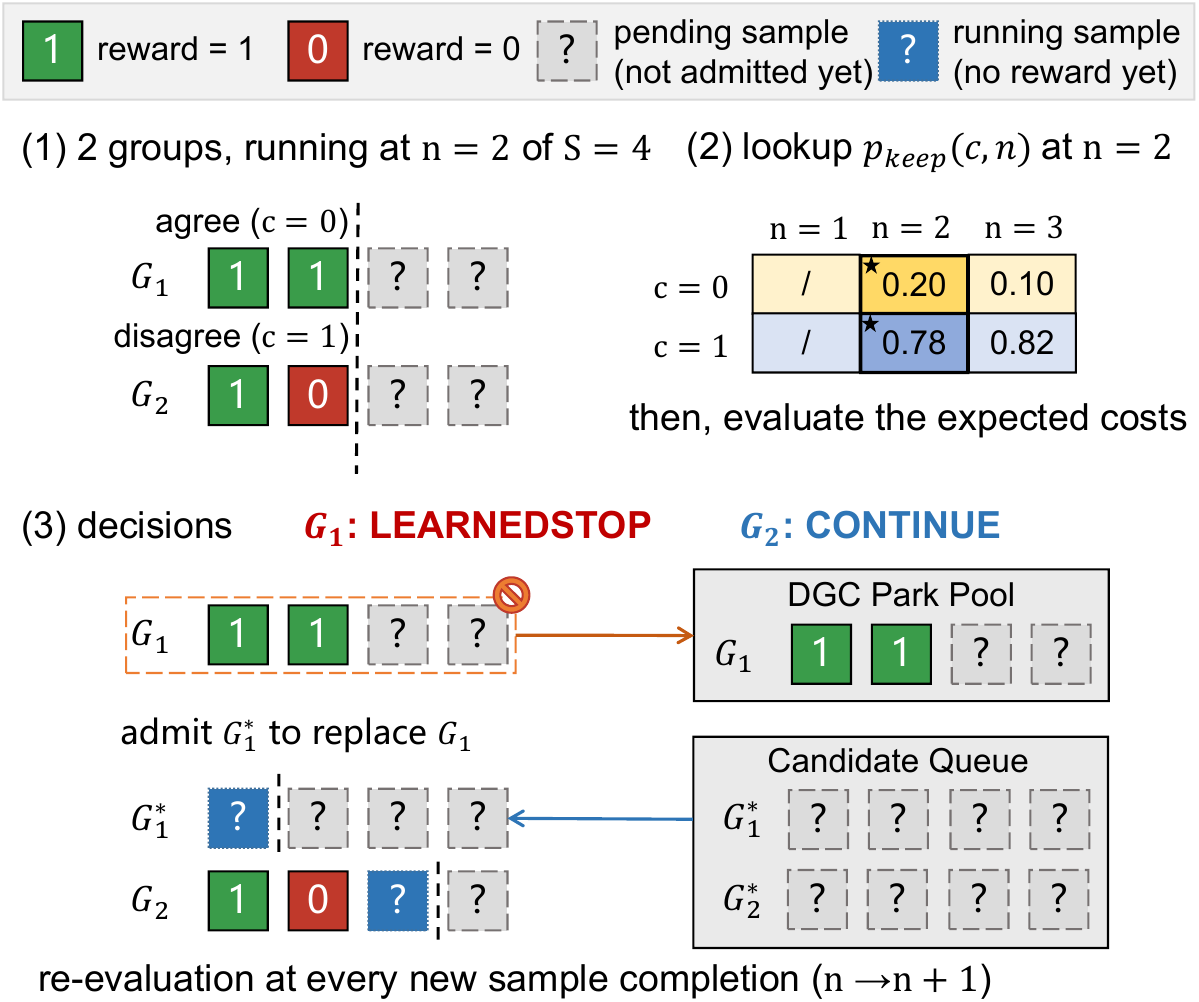}
\caption{\CSB's stop-vs-continue decision, contrasted for two groups at the same instant ($S{=}4$, $n{=}2$): (1) $G_1$'s completed samples agree ($c{=}0$) while $G_2$'s already disagree ($c{=}1$); (2) the calibration table returns a low $p_{\mathrm{keep}}$ for $G_1$ and a high one for $G_2$, and the cost-aware rule (Equation~\ref{eq:csb-threshold}) decides; (3) $G_1$ is stopped, parked in \DGC's pool and replaced by an oversampling candidate, while $G_2$ continues, re-evaluated at every new sample completion. Values are illustrative.}
\vspace{-0.1mm}
\label{fig:csb-flow}
\end{figure}

\CSB controls group-level continuation through two connected mechanisms: an online calibration table that turns partial observations into a calibrated keep probability, and a cost-aware rule that prices continuation against replacement. The division of responsibility is deliberate: \CSB releases the group's remaining slots at once, while admitting a replacement into them is \AAP's decision (\S\ref{sec:aap}), so a stop can be issued without knowing whether a replacement is available. For each candidate group, the rollout manager tracks a stable group identifier, number of completed samples, outstanding sample handles, partial rewards and filter signals, current verdict state, stop reason, and estimated continuation probability. In the \CSB path, a group's state transition is monotonic within a step:
\begin{equation*}
\textsc{Fresh}\rightarrow\textsc{Active}\rightarrow
\{\textsc{Keep},\textsc{Drop},\textsc{LearnedStop}\}.
\end{equation*}
On a deterministic filter failure, the group transitions directly to \textsc{Drop}; on a warm calibrated decision it may transition to \textsc{LearnedStop}, whose terminal handling \S\ref{sec:deferred-completion} replaces with a recoverable lifecycle; otherwise it continues generating its remaining samples until its ordinary filter verdict. Figure~\ref{fig:csb-flow} illustrates a concrete instance ($S{=}4$, $n{=}2$): $G_1$'s completed samples agree, so the lookup returns a low keep probability and the cost-aware rule (\S\ref{sec:csb-stop}) stops it without dispatching its pending samples, parks it in \DGC's pool in case the stop was a misfire, and lets a fresh oversampling candidate fill the freed slot, while $G_2$'s disagreeing samples return a high probability and let it continue, re-evaluated at each new completion ($n\to n{+}1$).

\subsubsection{Calibrated continuation probability}

After $n<S$ samples of a group finish, which probe sampling makes observable as $n$ advances (\S\ref{sec:probe}), \CSB maps the observable state, including completion count and partial reward and filter signals, to $p$, a calibrated estimate of the probability that the fully completed group will be kept. Both halves of this name are deliberate: $p$ comes from a calibration table maintained online rather than an explicit uncertainty model, and a stop frees slots for a replacement within the same batch rather than deferring the replacement to the next.

Calibration stores exponentially decayed \textsc{Keep} and \textsc{Drop} counts for each partial-state bucket $(c,n)$, where $c\in\{0,1\}$ marks whether the group's completed samples already disagree (on reward or filter outcome) and $n$ is the completed count; buckets are updated once per training step rather than once per completed group, with observations buffered and one decay plus one batched count update applied at the step boundary (\texttt{end\_step()}). This granularity is a deliberate design choice. Updating once per completed group would make a stream of unit observations approach an effective count of
\begin{equation}
\sum_{i=0}^{\infty}\lambda^i=\frac{1}{1-\lambda},
\end{equation}
which for decay $\lambda=0.85$ is $6.667$, and which may never reach the configured warm-up threshold $\tau_{\mathrm{warm}}$. Batching updates at the step boundary instead lets the effective history advance with training progress, so the count eventually clears warm-up. Learned stopping and probability-driven admission stay disabled until then; before warm-up, only deterministic filter rules apply. Fully completed groups and deterministic outcomes can update calibration, but a learned stop cannot be labeled \textsc{Drop}: its counterfactual verdict is unknown, so feeding it back would bias the table. The \textbf{Calibrated Continuation Probability Update} (Algorithm~\ref{alg:calib}) formalizes this update; a bucket that has not yet accumulated enough weight falls back first to the same $c$ at the nearest observed $n$, then to the global rate $p_0$.

\begin{algorithm}[t]
\caption{Calibrated Continuation Probability Update}
\label{alg:calib}
\begin{algorithmic}[1]
\Require step outcomes $\mathcal{O}=\{(c_i,n_i,y_i)\}$
\Require decay $\lambda$, warm-up $\tau_{\mathrm{warm}}$
\ForAll{bucket $(c,n)$}
  \State $\mathrm{hit}[c,n]\gets\lambda\cdot\mathrm{hit}[c,n]$
  \State $\mathrm{tot}[c,n]\gets\lambda\cdot\mathrm{tot}[c,n]$
\EndFor
\State $n_{\mathrm{keep}}\gets\lambda\cdot n_{\mathrm{keep}}$;\ $n_{\mathrm{seen}}\gets\lambda\cdot n_{\mathrm{seen}}$
\ForAll{$(c_i,n_i,y_i)\in\mathcal{O}$}
  \State $\mathrm{tot}[c_i,n_i]\mathrel{+}=1$;\ $\mathrm{hit}[c_i,n_i]\mathrel{+}=y_i$
  \State $n_{\mathrm{seen}}\mathrel{+}=1$;\ $n_{\mathrm{keep}}\mathrel{+}=y_i$
\EndFor
\ForAll{bucket $(c,n)$ with $\mathrm{tot}[c,n]>0$}
  \State $p_{\mathrm{keep}}(c,n)\gets \mathrm{hit}[c,n]/\mathrm{tot}[c,n]$
\EndFor
\State $p_0\gets n_{\mathrm{keep}}/\max(n_{\mathrm{seen}},1)$
\State \textsc{IsWarm}$\gets(n_{\mathrm{seen}}\ge\tau_{\mathrm{warm}})$
\State clear $\mathcal{O}$
\end{algorithmic}
\end{algorithm}

\subsubsection{Cost-aware stopping}
\label{sec:csb-stop}

\begin{table}[t]
\centering
\small
\setlength{\tabcolsep}{3pt}
\caption{Notation of the cost-aware stopping rule.}
\begin{tabularx}{\columnwidth}{@{}ll>{\raggedright\arraybackslash}X@{}}
\toprule
Role & Symbol & Meaning \\
\midrule
Belief & $p_{\mathrm{keep}}(c,n)$ & calibrated probability the completed group is kept. \\
Cost & $C_{\mathrm{rem}}$, $C_{\mathrm{repl}}$ & remaining-generation vs.\ replacement cost. \\
Pool state & $R$, $J$, $p_0$ & fresh candidates, concurrent stop seekers, base keep rate. \\
Threshold & $p^{*}(n,R,J)$ & stop when $p_{\mathrm{keep}} < p^{*}$. \\
\bottomrule
\end{tabularx}
\label{tab:csb-notation}
\vspace{-4mm}
\end{table}

\CSB decides when to stop a group by comparing expected cost. Stopping forgoes the group's remaining generation but must still supply one replacement; continuing pays for that generation and, should the group ultimately turn out to be a \textsc{Drop} (probability $1-p$), a replacement as well. Here $p$ is the group's current calibrated keep probability $p_{\mathrm{keep}}(c,n)$ (\S\ref{sec:csb}), $C_{\mathrm{rem}}$ the cost of finishing its remaining samples, and $C_{\mathrm{repl}}$ the expected cost of obtaining a replacement accepted group. The two expected costs are $\mathbb{E}[C\mid\textsc{LearnedStop}] = C_{\mathrm{repl}}$ and $\mathbb{E}[C\mid\textsc{Continue}] = C_{\mathrm{rem}} + (1-p)\,C_{\mathrm{repl}}$, so the learned rule prefers stopping when
\begin{equation}
p \;<\; \frac{C_{\mathrm{rem}}}{C_{\mathrm{repl}}}.
\label{eq:csb-threshold}
\end{equation}

A group is worth finishing only while its calibrated keep probability exceeds that ratio; this is an expected-cost decision, not an oracle guarantee.

A stop is useful, however, only if a replacement remains feasible. Let $R$ denote the number of unadmitted fresh candidate groups remaining in DAPO's oversampling pool (\S\ref{sec:background}), which is the rule's supply side, and $J$ the number of candidate groups simultaneously seeking a replacement at the current instant, which is its demand side. If each fresh candidate independently reaches acceptance with base probability $p_0$ (the global keep rate maintained by the calibrator, Algorithm~\ref{alg:calib}), then with $J{=}1$ the probability of obtaining a replacement is $1-(1-p_0)^{R}$, and in general \CSB evaluates the joint feasibility
\begin{equation}
P_{\mathrm{joint}}(R,p_0,J)=\Pr[\mathrm{Binomial}(R,p_0)\ge J].
\label{eq:joint-probability}
\end{equation}
Under a symmetric-duration assumption (every candidate is equally likely to run as long as any other, so both costs are expressed as fractions of one group's total sample count $S$), Equations~\ref{eq:csb-threshold} and~\ref{eq:joint-probability} combine into the closed-form threshold
\begin{equation}
p^{*}(n,R,J) = \underbrace{\frac{(S-n)\,p_0}{S}}_{\text{remaining work}}\cdot\;
\underbrace{\max\big(P_{\mathrm{joint}}(R,p_0,J),\,\epsilon\big)}_{\text{replacement feasibility}},
\label{eq:pstar-closed-form}
\end{equation}
and a learned stop triggers when $p_{\mathrm{keep}}(n) < p^{*}(n,R,J)$. The remaining-work factor scales the threshold with what the group has left to generate, so stopping is cheapest early and the threshold vanishes as $n\to S$; the feasibility factor tightens it when replacements are scarce, as the pool empties ($R\downarrow$) or demand rises ($J\uparrow$); $\epsilon$ is a floor that keeps the threshold from collapsing to zero.

The threshold requires a $J$ for every decision, yet stop decisions arrive concurrently across all active groups. Rather than solving a joint fixed point over them, \CSB computes $J$ directly at each evaluation: every other active group that already passes its own single-competitor ($J{=}1$) threshold is counted as a concurrent stop seeker (Algorithm~\ref{alg:stop}). This one-pass count makes the rule conservative under simultaneous stop demand without cross-group iteration.

\begin{algorithm}[t]
\caption{Cost-Aware Stopping Decision}
\label{alg:stop}
\begin{algorithmic}[1]
\Require $n$, $S$, calibrated $p_{\mathrm{keep}}(\cdot)$, base rate $p_0$, fresh candidates $R$, active groups $\mathcal{U}$
\Ensure $v\in\{\textsc{Continue},\textsc{LearnedStop}\}$
\If{$n<2$ \textbf{or} \textbf{not} \textsc{IsWarm} \textbf{or} resume-locked ($\ell_g$, \S\ref{sec:deferred-completion})}
  \State \Return \textsc{Continue}
\EndIf
\State $J\gets 1$
\ForAll{$g'\in\mathcal{U}$} \Comment{groups past $J{=}1$ threshold}
  \State $p^{*}_1\gets \frac{(S-n_{g'})p_0}{S}\max\big(1-(1-p_0)^{R},\epsilon\big)$
  \If{$p_{\mathrm{keep}}(n_{g'}) < p^{*}_1$} \State $J\gets J+1$ \EndIf
\EndFor
\State $p^{*}\gets\frac{(S-n)p_0}{S}\max\big(P_{\mathrm{joint}}(R,p_0,J),\epsilon\big)$
\If{$p_{\mathrm{keep}}(n) < p^{*}$}
  \State \Return \textsc{LearnedStop}
\Else
  \State \Return \textsc{Continue}
\EndIf
\end{algorithmic}
\end{algorithm}

\subsubsection{What the rule approximates}

Equations~\ref{eq:csb-threshold} and~\ref{eq:pstar-closed-form} price continuation against replacement in units of one group's own generation, whereas the full objective a barrier-bound scheduler faces also includes the delay added to the step's critical path ($C_{\mathrm{barrier}}$) and the cost of an extra admission batch when too few accepted groups arrive in time ($C_{\mathrm{shortfall}}$). Neither is observable when a stop must be decided: only the fraction of remaining generation landing on the critical path delays the barrier, and a shortfall is realized only once replacement demand outruns the pool, forcing a late, disproportionately expensive batch. \CSB substitutes one measurable surrogate for both, the replacement-feasibility factor $P_{\mathrm{joint}}(R,p_0,J)$ (Equation~\ref{eq:joint-probability}), which turns conservative exactly when replacements are scarce, backed by the safety guards of \S\ref{sec:csb-safety} that fall back to continuation when joint feasibility is low. The two equations are thus that objective specialized to a symmetric-duration, on-critical-path, replacement-feasible regime.

\subsubsection{Replacement on stop}

When \CSB stops a group, the scheduler cancels or suppresses its not-yet-started samples, records the work released, and frees the slot; running requests may not be preemptible at token granularity, so realized savings depend on stop timing and on which requests the backend can remove. What refills the slot, and whether any further candidate is admitted at all, is \AAP's decision (\S\ref{sec:aap}).

\subsubsection{Safety and fallback}
\label{sec:csb-safety}

Cold start disables both learned stopping and probability-driven admission, leaving only deterministic filter rules. Runtime guards also disable a learned stop when too few fresh candidates remain, joint replacement probability is below threshold, or the target batch may not fill, sacrificing savings to preserve progress toward the barrier. Telemetry records calibration counts, learned and deterministic stop counts, groups touched, backfills, admission rounds, shortfall waves, and per-step rollout and barrier durations, so that a measured change is attributable to a decision path rather than to an aggregate speedup.

\subsection{Deferred Group Completion}
\label{sec:deferred-completion}

\CSB's learned stop saves current-step resources but may discard a true-KEEP group. \DGC remedies this error by replacing \emph{cancellation} with \emph{parking}: a stopped group keeps every sample it has already finished, and is resumed later from where it stopped instead of being discarded. 

\subsubsection{Parking in place of cancellation}

Parking releases a group's pending GPU slots while retaining every already-finished sample, its tokens, reward, and policy-version metadata, the group's identity, and how many samples remain. A sample still in flight when its group is parked is not partially preserved as training data, and resume later regenerates only the still-missing samples. A group enters training only after reaching $S$ samples and receiving its final verdict, so \DGC parks and resumes a \emph{GRPO group}, never an incomplete trajectory used directly as a training sample.

Parking is allowed only for probabilistic \texttt{learned\_stop} results. Deterministically invalid conditions (e.g., reward exceptions, inconsistent predictions, or truncation-ratio breaches) are filtered out, since restarting them cannot overturn an already decided outcome. The lifecycle of a \textsc{Parked} group can be described as a simple state machine: a qualifying \textsc{LearnedStop} transitions from \textsc{Active} to \textsc{Parked} as long as capacity remains under $M$, the maximum size of the deferred-group pool; a \textsc{Parked} group turns \textsc{Active} when resumed, or transitions to \textsc{Drop} due to aging, incompatibility, or a bounded-pool replacement rule; deterministic \textsc{Drop} states are excluded from this lifecycle entirely. The scheduler manages the pool, and thus $M$, per rollout step rather than per SGLang engine, so $M$ bounds \textsc{Parked} groups across a whole step and does not scale with engine count.

\begin{figure}[t]
\centering
\includegraphics[width=\linewidth]{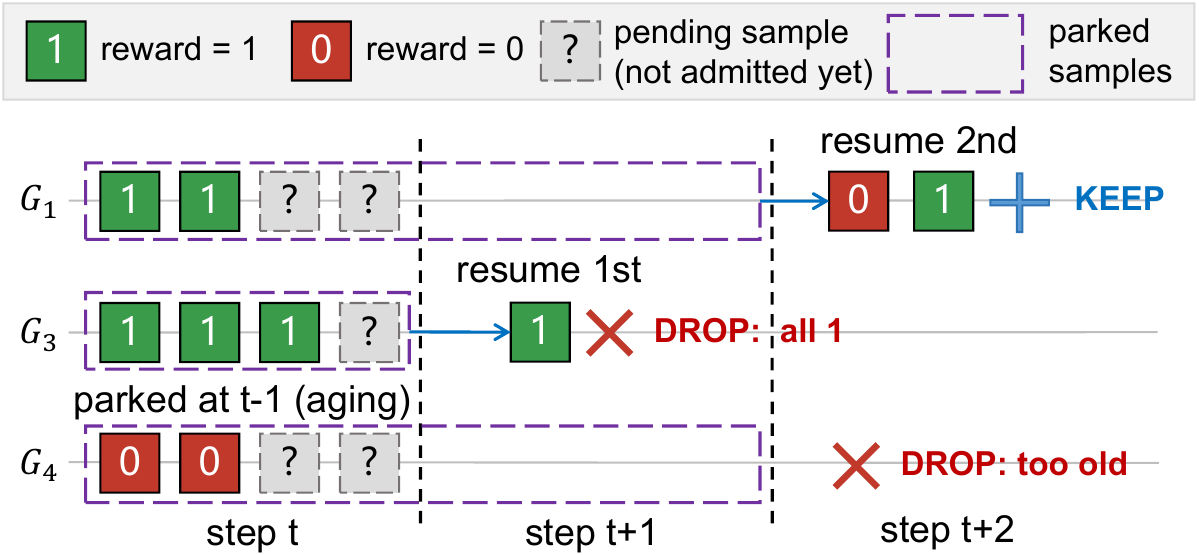}
\caption{\DGC's park/resume lifecycle for the three groups stopped in Figure~\ref{fig:csb-flow}: $G_3$ confirms its stop, $G_1$ is a \textsc{Keep} that discarding would have lost, and $G_4$ ages out. Values are illustrative.}
\vspace{-0.1mm}
\label{fig:dgc-states}
\end{figure}

Figure~\ref{fig:dgc-states} extends the scenario of Figure~\ref{fig:csb-flow} to illustrate why the state that parking retains is worth retaining: all three groups were halted for the same cause, namely that their completed samples agreed, yet their outcomes diverge. $G_3$, the most complete, is resumed first, completes with all samples agreeing, and is then dropped, confirming the stop; $G_1$ waits one step and, once resumed, yields disagreeing samples, a \textsc{Keep} that discarding outright would have lost permanently; $G_4$, parked a step earlier, is never revisited and eventually ages out. 

\subsubsection{Resume order and pool bounds}

A bounded priority queue ranks eligible groups by completion progress and uses age as a starvation guard, so the most-completed parked group is resumed first and no group waits indefinitely. Resumption validates model and tokenizer compatibility and continues from the retained KV cache when the runtime still holds it. Resume dispatch reuses the group's persisted probe-duration priority $\mu_g$ (\S\ref{sec:probe}) instead of a fresh flat-priority ticket, keeping resumed work consistently ordered with \AAP's pipelining. Algorithm~\ref{alg:park} formalizes \textbf{Deferred Group Completion}'s park lifecycle through three procedures: \textsc{TryPark}, \textsc{PopBestAndResume}, and the per-step aging pass \textsc{EndStepAge}.

\begin{algorithm}[t]
\caption{Deferred Group Completion}
\label{alg:park}
\begin{algorithmic}[1]
\Require pool $\mathcal{P}$, capacity $M$, age limit $T_{\mathrm{age}}$
\Procedure{TryPark}{$g$} \Comment{on \textsc{LearnedStop}}
  \If{$M\le 0$ \textbf{or} $|\mathcal{P}|\ge M$}
    \State \Return \textsc{False} \Comment{discard $g$}
  \EndIf
  \State add $g$ to $\mathcal{P}$; $r_g\gets0$; $\ell_g\gets\textsc{False}$; $\mathrm{age}_g\gets0$
  \State \Return \textsc{True}
\EndProcedure
\Statex
\Procedure{PopBestAndResume}{}
  \If{$\mathcal{P}=\emptyset$} \State \Return \textsc{None} \EndIf
  \State $g\gets\arg\max_{g'\in\mathcal{P}} n_{g'}$ \Comment{most complete first}
  \State remove $g$ from $\mathcal{P}$;\ $r_g\gets r_g+1$
  \If{$r_g > C_{\max}$} \State $\ell_g\gets\textsc{True}$ \Comment{lock: no more stops} \EndIf
  \State ticket $\gets \mu_g$ if known, else FIFO
  \State dispatch remaining $S-n_g$ samples with ticket
  \State \Return $g$
\EndProcedure
\Statex
\Procedure{EndStepAge}{}
  \ForAll{$g\in\mathcal{P}$}
    \State $\mathrm{age}_g\mathrel{+}=1$
    \If{$\mathrm{age}_g > T_{\mathrm{age}}$} \State remove $g$ from $\mathcal{P}$ \EndIf
  \EndFor
\EndProcedure
\end{algorithmic}
\end{algorithm}

\subsubsection{Bounded resume cycles}

A resumed group already satisfies \CSB's minimum-sample precondition for a learned-stop evaluation, since it was parked precisely because that precondition was met. Left unguarded, this creates an oscillation risk: if the first sample to complete after resume again looks unfavorable, \CSB would stop the group a second time, parking it again, and this could repeat indefinitely without ever reaching a real verdict. \DGC bounds this directly with a per-group resume-cycle counter $r_g$: once a group has been resumed more than $C_{\max}$ times (zero by default, that is, after its first resume), it is resume-locked ($\ell_g$) and permanently exempted from further learned-stop evaluation, so it must run the rest of its samples to a deterministic or full-completion verdict. This trades a small amount of additional, but bounded, generation for a guarantee that every parked group eventually reaches a real outcome. This guard is not merely precautionary: in an early deployment without it, groups cycled between \textsc{Parked} and \textsc{Active} about $8.4$ times on average before reaching a verdict, re-dispatching and then cancelling their remaining samples each time, and step time rose $55\%$ instead of falling. Bounding the cycle is what turned deferral from a net loss into the $4.46\%$ gross cost of \S\ref{sec:component-breakdown}.

\subsubsection{Cost and correctness of deferral}

\DGC is not merely extra overhead: a paused group can draw on unused GPU capacity in a subsequent step instead of contending with the step that caused it to pause; prefix and KV-cache reuse can prevent re-generating tokens that were already emitted; and recovery lowers the number of new candidates that must be fetched from the pool. However, the reported savings never rely on a cache hit, and when a group resumes after a policy update, it may include samples generated under an earlier policy version than those produced anew at resume time. The scheduler therefore logs the policy version for every sample and excludes a stale retained prefix from the policy-gradient loss. \S\ref{sec:discussion} describes how synchronous semantics are maintained when deferred completion extends across a policy update.

\subsection{Adaptive Admission and Pipelining}
\label{sec:aap}

\CSB and \DGC decide \emph{which} group-level computation is worth keeping. \AAP decides how much of it the runtime admits at any moment and in what order it executes, across every path a candidate group takes through \SystemName, freshly admitted candidates and groups \DGC resumes alike (\S\ref{sec:deferred-completion}). It combines two orthogonal sub-mechanisms.

\subsubsection{Ordering from early completions}

The duration signal that probe sampling carries (\S\ref{sec:probe}) is derived online~\cite{he2025history}: it becomes available once the first samples of a group complete and is recomputed as further samples finish, so the ordering tracks measured duration rather than a prediction made once and held. Groups measured to run long are scheduled with higher priority~\cite{xu2026war}. This ordering applies uniformly across \SystemName's rollout flow: it governs freshly admitted candidates and, because \DGC persists a parked group's current priority, also governs a resumed group's remaining samples. \AAP only consumes this signal and answers only which admitted sample runs next, never \CSB's continuation decision, so disabling it falls back to FIFO.

\subsubsection{Supply-aware streaming admission}

The second sub-mechanism replaces batch-wise admission, in which a conventional dynamic-sampling rollout admits its oversampled candidates as one batch and drains it before admitting more, with a continuous supply-aware rule. It has two parts.

\emph{Streaming (pipelining).} \AAP refills each slot immediately once a group's fate is known, whether by \CSB, a deterministic filter check, or \DGC, drawing from the candidate pool or \DGC's park pool, rather than waiting for the rest of the batch to drain. This keeps the fixed-capacity pipeline saturated for the whole step instead of oscillating between a full batch and a draining one~\cite{wang2025seamlessflow}.

\emph{Supply-aware admission count.} \AAP admits another candidate only while the expected supply of accepted groups still falls short of the step's target batch, in the vocabulary of dynamic sampling~\cite{yu2026dapo}:
\[
\bigl|\{\text{accepted so far}\}\bigr|
\;+\!\!\sum_{g\,\in\,\text{undecided}}\!\! p_{\mathrm{keep}}(g)
\;<\; B,
\]
where $B$ is the number of accepted groups the training batch requires and $p_{\mathrm{keep}}(g)$ is exactly the belief signal \CSB already maintains for every in-flight group. Before the calibrator has accumulated enough evidence to rely on that signal, \AAP falls back to a pessimistic bound that counts every undecided group as accepted, and the operator-set oversampling factor caps how many groups may be admitted in total. The rule yields an admission count only, leaving each group's continuation decision to \CSB.

\subsection{Unified Control Loop}
\label{sec:control-loop}
At each scheduling event the integrated loop applies \CSB's deterministic checks and calibrated continuation decisions as samples complete, allows \DGC to park eligible learned stops or to resume the most advanced parked group when a candidate is needed, and allows \AAP to decide whether another candidate should be admitted and in what order admitted work runs. The training barrier stays at $B$ complete accepted groups throughout, and each mechanism degrades along its own fallback path (\S\ref{sec:csb} and \S\ref{sec:aap}): before warm-up \CSB applies only deterministic rules and \AAP admits on the pessimistic bound, and if \DGC's pool is full or a group ages past $T_{\mathrm{age}}$, deferral reverts to the terminal discard it replaced.

We implement \SystemName on top of slime v0.2.4~\cite{slime_github}, as a set of pluggable scheduling and control components: \CSB, \DGC, and \AAP attach to slime's existing sample-completion and admission hooks, so slime's own rollout, generation backend, and training loop are unmodified when \SystemName is disabled.

\section{Evaluation}
\label{sec:evaluation}



This section presents end-to-end online evaluation results (\S\ref{sec:online-results}), performance breakdown (\S\ref{sec:component-breakdown}), the analysis of \CSB and \DGC on true-KEEP recall (\S\ref{sec:trace-replay}), and
long-horizon stability and cross-model generalization (\S\ref{sec:extended}).

\subsection{Setup}
\label{sec:methodology}

\cparagraph{Testbed} Our testbed comprises 2 nodes, each with 8 NVIDIA H20 GPUs, dual Intel Xeon Platinum 8558 processors (96 cores), 2.2TB DRAM and 25.6TB NVMe storage; we deploy \SystemName colocated, with GPU count and parallelization chosen per model.

\begin{table}[t]
\caption{Model and workload configurations.}
\label{tab:workloads}
\centering
\small
\setlength{\tabcolsep}{4pt}
\begin{tabular}{@{}lrrr@{}}
\toprule
Metric & Qwen3.5- & \multicolumn{2}{c}{Qwen3.5-4B} \\
       & 35B-A3B  & phase 1 & phase 2 \\
\midrule
Workload                  & long CoT & instruction & instruction \\
Target Groups/Step ($B$)  & 16       & 128         & 128         \\
Samples/Group ($S$)       & 16       & 16          & 16          \\
Response Budget           & 100K     & 16K         & 32K         \\
Rollout Steps             & 20       & 220         & 135         \\
Total GPUs                & 8        & 16          & 16          \\
\bottomrule
\end{tabular}
\end{table}

\cparagraph{Models and Workloads} We evaluate two models of different architecture and scale on the same math-only data drawn from OpenMathReasoning~\cite{moshkov2025aimo2} and DAPO-Math-17k~\cite{yu2026dapo}, trained toward different response styles (Table~\ref{tab:workloads}). Qwen3.5-35B-A3B~\cite{qwen3technicalreport} trains for long chain-of-thought reasoning; the dense Qwen3.5-4B~\cite{qwen3technicalreport} answers directly in an instruction-following style, in two phases, first a short-decode workload, then a longer budget that exercises \CSB's calibration under a genuine distribution shift (\S\ref{sec:extended}).

All rollout-time, step-time, and throughput results come from slime's logged per-step metrics, averaged over each run's steps (Table~\ref{tab:workloads}). Each final checkpoint is also scored on a math suite and an instruction-following suite (\S\ref{sec:online-results}), so speed and model quality are compared together.

\cparagraph{Settings} We use DAPO's dynamic-sampling GRPO formulation (\S\ref{sec:background}); $B$, $S$, and the oversampling size ($3\times B$, the best ratio found when tuning our production deployment) are fixed across every configuration and baseline, so differences in rollout time and generation stem only from the mechanism under test. The controller's own parameters are the decay $\lambda=0.85$ and warm-up $\tau_{\mathrm{warm}}=40$ effective observations of \S\ref{sec:csb}, a small feasibility floor $\epsilon$, and for \DGC the pool capacity $M$, age limit $T_{\mathrm{age}}$ and resume-cycle bound $C_{\max}$, set to $M{=}16$, $T_{\mathrm{age}}{=}3$ steps and $C_{\max}{=}0$ on the 35B workload and $M{=}64$ on the 4B run (\S\ref{sec:extended}). Training uses Adam with DAPO's asymmetric clip ($\epsilon_{\mathrm{low}}{=}0.2$, $\epsilon_{\mathrm{high}}{=}0.28$) and per-token loss without a KL penalty; rollout uses temperature $1.0$ and each configuration's response budget.

\cparagraph{Baselines} We compare against two external baselines, an internal ablation, and an oracle ceiling:
\begin{itemize}[leftmargin=1.2em,topsep=0pt,partopsep=0pt]
\item \emph{Slime}: slime's default rollout configuration, with no additional stopping, recovery, or admission control.
\item \emph{Seer}: probe-based reordering at our group-and-sample granularity~\cite{qin2026seer}, changing only the order of already admitted work.
\item \emph{\SystemName\ (\CSB\ only)}: \CSB over probe ordering without \DGC and \AAP, isolating \CSB's own speed/recall trade-off.
\item \emph{Oracle \SystemName}: \CSB's calibrated probability replaced by the counterfactual verdict from trace replay (\S\ref{sec:trace-replay}), the ceiling of a perfect stop.
\end{itemize}

\begin{table*}[t!]
\caption{End-to-end comparison over 20 steps ($B=16$, $S=16$). Lower is better for times, higher for throughput; bold is the full \SystemName configuration with its change versus Slime. Oracle \SystemName is a trace-replay bound.}
\label{tab:online-arc}
\centering
\small
\begin{tabular}{lrrrrr}
\toprule
Metric & Slime & Seer & \SystemName (\CSB\ only) & \SystemName & Oracle \SystemName \\
\midrule
Rollout time (s) & 194,999 & 193,372 & 119,152 & \textbf{122,087 ($-37.39\%$)} & 81,221 \\
Step time (s) & 219,889 & 219,330 & 146,286 & \textbf{146,538 ($-33.36\%$)} & 106,111 \\
GPU rollout throughput (tokens/GPU/s) & 95.80 & 99.70 & 171.90 & \textbf{153.69 ($+60.43\%$)} & 236.57 \\
Effective GRPO throughput (groups/h) & 5.24 & 5.25 & 7.87 & \textbf{7.86 ($+50.1\%$)} & 10.86 \\
\bottomrule
\end{tabular}
\end{table*}

\subsection{End-to-End Performance}
\label{sec:online-results}

\begin{table}[t!]
\caption{Downstream accuracy of each checkpoint: math and instruction-following/reasoning benchmark groups.}
\label{tab:model-reward}
\centering
\small
\begin{tabular}{lrrr}
\toprule
Benchmark & Slime & \SystemName (\CSB\ only) & \SystemName \\
\midrule
\multicolumn{4}{l}{\textit{Math}} \\
MATH-500 & 98.05\% & 97.70\% & 97.75\% \\
AIME24 & 91.46\% & 91.04\% & 91.25\% \\
AIME25 & 90.00\% & 89.17\% & 90.42\% \\
AIME26 & 91.25\% & 91.25\% & 89.38\% \\
HMMT-Feb25 & 86.46\% & 85.63\% & 85.83\% \\
HMMT-Feb26 & 76.89\% & 76.70\% & 77.84\% \\
\quad mean & 89.02\% & 88.58\% & 88.74\% \\
\midrule
\multicolumn{4}{l}{\textit{Reasoning}} \\
MuSR & 69.18\% & 67.99\% & 68.78\% \\
IFEval & 92.61\% & 90.94\% & 91.87\% \\
IFBench & 54.67\% & 60.67\% & 59.33\% \\
\quad mean & 72.15\% & 73.20\% & 73.33\% \\
\bottomrule
\end{tabular}
\end{table}

Table~\ref{tab:online-arc} compares all configurations over the same 20-step, $B{=}16$, $S{=}16$ workload: every run targets the same 320 accepted groups, so the configurations differ only in the time and generation required to reach that target. We examine step time and throughput, then downstream accuracy, then the distribution of trajectories reaching training.

\subsubsection{Rollout time}

Rollout time falls $37.39\%$ and step time $33.36\%$ under the full \SystemName configuration. Seer's reordering is nearly neutral over the same window ($-0.83\%$ rollout time, $+0.25\%$ effective throughput), against the $38.9\%$ that \CSB recovers on the same workload. \SystemName\ (\CSB\ only) is marginally faster on every row, the difference being \DGC's residual $0.17\%$ step-time cost (\S\ref{sec:component-breakdown}) paid for recovering the groups \CSB alone discards.

\subsubsection{Effective token throughput}

GPU rollout throughput (tokens/GPU/s) rises $60.43\%$ for \SystemName and $79.4\%$ for \SystemName\ (\CSB\ only). Its numerator counts only the tokens of samples that reach the training batch, so it is a rate of \emph{useful} generation: the $150.1$M useful tokens of the full \SystemName configuration fall below the $272.0$M it generates in total (Table~\ref{tab:dgc-composition}). Slime generates $445.0$M tokens to deliver $149.4$M, a useful fraction of $33.6\%$, the over-sampling surplus that DAPO's filter discards by design; \SystemName's $55.2\%$ cuts the discarded share by $21.6$ percentage points. \emph{Effective GRPO throughput} (accepted groups per training hour) rises $50.1\%$, measuring the same delivery at group granularity and against full step time.

\subsubsection{Downstream accuracy}

Speed and throughput alone are uninformative about policy quality. Table~\ref{tab:model-reward} scores each checkpoint on six math~\cite{lightman2023letsverify,balunovic2025matharena} and three instruction-following, reasoning benchmarks~\cite{sprague2024musr,zhou2023ifeval,pyatkin2025ifbench}: all configurations stay within $1.87$pp on any single math benchmark and $0.44$pp on the math mean, inside the noise expected at this scale, so neither \CSB's stops nor \DGC's recovery measurably changes what the policy learns. The full configuration's math mean ($88.74\%$) sits between Slime's and \SystemName\ (\CSB\ only)'s, consistent with \DGC recovering some of the groups \CSB alone drops; on instruction following \SystemName\ (\CSB\ only) averages $1.05$pp \emph{higher} than Slime ($+6.00$pp on IFBench), which we attribute to sampling noise, though it rules out systematic degradation. To rule out a run that diverges mid-way and recovers by the end, we tracked training-time \texttt{task\_score} step by step: both \SystemName configurations deviate from Slime's cumulative mean by no more than a single run's own step-to-step noise ($\pm0.11$), without growing over the run.

\begin{table}[t]
\caption{Per-sample response length within the groups accepted into training.}
\label{tab:length-dist}
\centering
\small
\begin{tabular}{lrrr}
\toprule
Statistic (tokens) & Slime & \SystemName (\CSB\ only) & \SystemName \\
\midrule
Mean & 27{,}582 & 29{,}772 & 27{,}485 \\
P50 & 16{,}403 & 19{,}774 & 17{,}746 \\
P90 & 71{,}168 & 72{,}378 & 70{,}062 \\
P99 & 96{,}456 & 97{,}174 & 94{,}978 \\
Truncation rate & $1.86\%$ & $2.48\%$ & $1.80\%$ \\
\bottomrule
\end{tabular}
\end{table}

\subsubsection{Response-length distribution}

Table~\ref{tab:length-dist} compares per-sample response lengths within the accepted groups, where the two \SystemName configurations differ sharply. \CSB alone shifts the accepted population measurably \emph{longer} (mean $+7.9\%$ and median $+20.5\%$ over Slime), because groups whose short samples finish early are confirmed and admitted quickly, leaving a larger share of the fixed training batch to long-running groups. The full configuration returns the mean to within $0.4\%$ of Slime, holds the upper percentiles to within $1.6\%$ (P90 $-1.6\%$, P99 $-1.5\%$), and narrows the median from $+20.5\%$ to $+8.2\%$: \DGC restores the shape of the training distribution, not merely the $98.9\%$ of recovered groups (\S\ref{sec:trace-replay}).

\subsection{Component Breakdown}
\label{sec:component-breakdown}

Table~\ref{tab:breakdown} attributes the end-to-end speedup, each configuration adding one mechanism on top of the last; all step-time figures include the scheduler's control overhead, microseconds to low milliseconds per decision. \CSB's share rests on step-batched rather than per-group calibration (\S\ref{sec:csb}): per-group decay leaves the effective observation count below warm-up, so the admission rule stays inactive and the run draws nearly the whole candidate pool, 968 against the 960 that $3\times B$ over 20 steps implies, while per-step decay activates it and lowers the count to 650, the configuration supplying \SystemName\ (\CSB\ only) in Table~\ref{tab:dgc-composition}, which also eliminates every shortfall wave and issues far fewer backfills (Table~\ref{tab:calib-granularity}). \DGC (Algorithm~\ref{alg:park}) is the only mechanism that costs step time, $4.46\%$ over \CSB alone, because a parked group holds host-side token state plus the KV cache of its prefix until resumed; it nonetheless generates fewer tokens, a resumed group regenerating only its remaining $S-n$ samples with no loss in the \texttt{task\_score} proxy (Table~\ref{tab:dgc-composition}). \AAP recovers most of that cost, taking the full configuration to $1.60\times$ rollout time and $1.50\times$ step time at unchanged $98.9\%$ recall.

\begin{table}[t]
\caption{\SystemName's performance improvement breakdown (higher $\times$ is faster).}
\label{tab:breakdown}
\centering
\small
\begin{tabular}{lccl}
\toprule
Configuration & Rollout speedup & Step speedup & Recall \\
\midrule
Slime & $1.00\times$ & $1.00\times$ & --- \\
w/ \CSB & $1.64\times$ & $1.50\times$ & $80.6\%$ \\
w/ \DGC & $1.54\times$ & $1.44\times$ & $98.9\%$ \\
w/ \AAP & $1.60\times$ & $1.50\times$ & $98.9\%$ \\
\bottomrule
\end{tabular}
\end{table}

\begin{table}[t]
\caption{\CSB's calibration update granularity.}
\label{tab:calib-granularity}
\centering
\small
\begin{tabular}{lrr}
\toprule
& Per-group updates & Per-step updates (used) \\
\midrule
Groups touched & 968 & 650 \\
Backfills issued & 626 & 274 \\
Shortfall waves & 11 & 0 \\
\bottomrule
\end{tabular}
\end{table}

\begin{table}[t]
\caption{\DGC's effect on rollout composition. Percentages are the change against Slime.}
\label{tab:dgc-composition}
\centering
\small
\setlength{\tabcolsep}{3pt}
\begin{tabularx}{\columnwidth}{@{}Xc>{\raggedleft\arraybackslash}p{1.5cm}>{\raggedleft\arraybackslash}p{1.5cm}@{}}
\toprule
Metric & Slime & \CSB\ only & \SystemName \\
\midrule
Fresh candidate groups & 960 & 650  & \textbf{587} \\
Tokens generated & 445.0M & 331.5M  & \textbf{272.0M} \\
Useful token fraction & $33.6\%$ & $49.4\%$ & \textbf{$55.2\%$} \\
\texttt{task\_score} & 0.334 & 0.318 & \textbf{0.363} \\
\bottomrule
\end{tabularx}
\end{table}

\subsection{Trace-Replay Analysis: \DGC's Capacity Trade-off}
\label{sec:trace-replay}

We replay two production traces, recorded during math-reasoning training of Qwen3.5-35B-A3B on \emph{Easy} and \emph{Hard}, two datasets obtained by splitting the math training set by problem difficulty and running reinforcement learning on each independently. Each spans 88 or 71 steps at 256 target groups, far more than the online run of \S\ref{sec:online-results}. We replay each trace twice: once as Slime, letting every group run to completion to reconstruct ground truth, and once as \CSB, with its decision policy advanced causally within the replay; stop decisions are then scored against that ground truth. The replay charges a resumed group only the prorated remainder of its stopped samples, matching prefix reuse, and keeps the pool across steps; under the conservative alternative of charging full regeneration, the $M{=}0$ point reproduces the online \CSB-only run exactly, so the parking logic introduces no bias.

\begin{figure}[t]
\centering
\includegraphics[width=0.72\linewidth]{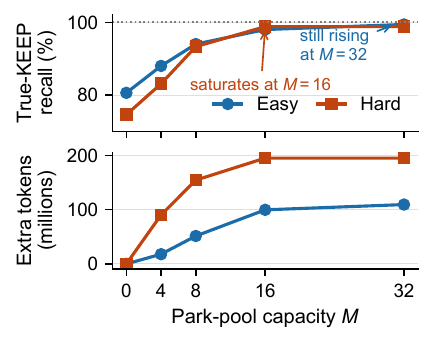}
\caption{\DGC capacity sweep on the two recorded traces: recall and extra tokens vs.\ park-pool capacity $M$; $M{=}0$ is \CSB with \DGC disabled.}
\vspace{-4mm}
\label{fig:park-simulation}
\end{figure}

Sweeping the park-pool capacity $M\in\{0,4,8,16,32\}$, where $M{=}0$ is \CSB with \DGC disabled (Figure~\ref{fig:park-simulation}), true-KEEP recall starts at $80.6\%$ on Easy and $74.6\%$ on Hard and rises steeply up to $M{=}16$ on both. Beyond it Easy still gains $1.4$ points ($98.0\%\rightarrow99.4\%$), while Hard is \emph{identical} at $M{=}16$ and $M{=}32$, its peak concurrent occupancy never exceeding $M{=}16$. Easy's $98.0\%$ at $M{=}16$ is close to the $98.9\%$ measured online (\S\ref{sec:component-breakdown}). The recovery costs extra tokens, which grow monotonically with $M$ up to each trace's saturation point.

\subsection{Long-Horizon Adaptation}
\label{sec:extended}

\begin{figure}[t!]
\centering
\includegraphics[width=0.82\linewidth]{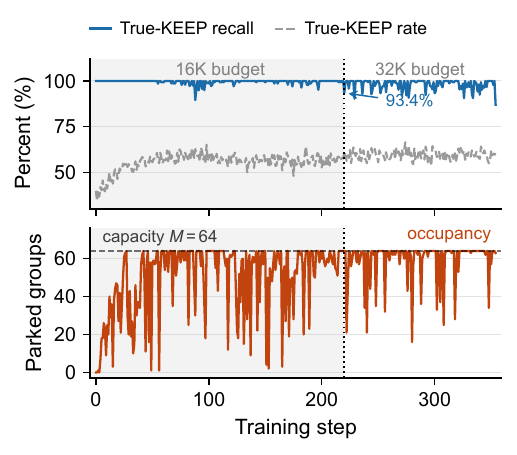}
\caption{The 355-step Qwen3.5-4B run. Dashed line in the top panel is the workload's own true-KEEP recall and rate, which rises after the budget change; the bottom panel shows park-pool occupancy against its capacity.}
\vspace{-0.1mm}
\label{fig:long-horizon}
\end{figure}

We next examine whether \SystemName remains effective over a long horizon with a changing rollout budget, and how the park-pool capacity should be provisioned.

\cparagraph{Long-horizon adaptation under a changing budget} We run a 355-step Qwen3.5-4B experiment with a $16$K rollout budget for the first 220 steps and $32$K for the remaining 135 steps. Table~\ref{tab:4b} shows \SystemName reducing makespan by $63.2\%$ and $52.6\%$ in the two phases while holding $99.6\%$ and $98.0\%$ true-KEEP recall, against $88.5\%$ and $84.5\%$ for \CSB alone.

\begin{table}[t]
\caption{Qwen3.5-4B: two consecutive phases of one run, $B=128$, $S=16$. Recall is true-KEEP recall; \CSB only is \SystemName with \DGC disabled.}
\label{tab:4b}
\centering
\small
\begin{tabular}{lrrr}
\toprule
& Slime & \SystemName (\CSB\ only) & \SystemName \\
\midrule
\multicolumn{4}{l}{\textit{Phase 1: 16K budget, 220 steps}} \\
Makespan (h) & 157.3 & 71.8 & \textbf{57.9} \\
Reduction & --- & $54.4\%$ & \textbf{$63.2\%$} \\
Recall & $100.0\%$ & $88.5\%$ & \textbf{$99.6\%$} \\
\midrule
\multicolumn{4}{l}{\textit{Phase 2: 32K budget, 135 steps}} \\
Makespan (h) & 112.3 & 66.3 & \textbf{53.2} \\
Reduction & --- & $41.0\%$ & \textbf{$52.6\%$} \\
Recall & $100.0\%$ & $84.5\%$ & \textbf{$98.0\%$} \\
\bottomrule
\end{tabular}
\end{table}

The tighter $16$K budget exposes more recoverable waste: $10.5\%$ of samples hit the limit, and Slime wastes $53.1\%$ of generated tokens in phase~1 versus $37.0\%$ in phase~2. Across all steps, token-waste share correlates with \SystemName's makespan reduction at $r=+0.84$, so the benefit tracks recoverable computation, not response length. At the budget transition the calibrator converges to within $0.7\%$ of the new steady-state keep rate in 9 steps without reset, and a warm-started phase-2 run differs from the continuous run by only $0.4\%$.

\begin{table}[tb]
\caption{Park-pool capacity scan on Qwen3.5-4B (phase 1). ``Refused'' counts learned stops denied a park slot because the pool was full.}
\label{tab:park-cap}
\centering
\small
\begin{tabular}{rrrr}
\toprule
$M$ & Recall & Refused \\
\midrule
16 (default)  & $92.9\%$ & 9859 \\
32  & $95.3\%$ & 6248 \\
64 (used here) & $\mathbf{99.6\%}$ & 476 \\
256 & $100.0\%$ & 0 \\
\bottomrule
\end{tabular}
\end{table}

\cparagraph{Recovery capacity is workload-dependent} The same run fixes a practical \DGC parameter, the park-pool capacity $M$ (Table~\ref{tab:park-cap}). This workload produces roughly 60 learned stops per step, so the $M{=}16$ of the breakdown is permanently saturated: it refuses $9859$ learned drops, $2.4\times$ the $4111$ it parks, and capacity rather than the stopping rule caps recall at $92.9\%$. $M{=}64$, used throughout this run, reaches $99.6\%$ and refuses only $476$ of $\sim$$14$k learned stops ($3.4\%$); $M{=}256$ closes the remaining gap. Conversely, the Hard trace of \S\ref{sec:trace-replay} never fills even $M{=}16$. $M$ should therefore be provisioned according to the workload's stopping rate: insufficient capacity turns recoverable stops into terminal discards, while excess capacity yields little additional benefit. Determining it online is left to future work.
\section{Discussion}
\label{sec:discussion}

\cparagraph{Scope of intervention} \SystemName leaves the training barrier and the group-shaped training interface intact, changing only which groups fill the batch and how their samples run. The one departure from strict per-sample synchrony is a parked group whose completion spans a policy update. This has not produced a stale prefix in either online \DGC deployment: cancellation aborts an in-flight sample at the client only, no parked candidate carried a non-zero pre-resume length, and every resumed sample is generated in full under the resume-time policy. At group granularity the question is bounded rather than excluded: a parked group ages out after $T_{\mathrm{age}}$ steps, capping the version gap, and must again supply all $S$ samples before any reach the update, so no partially generated trajectory enters policy optimization. The version-tracking and loss-masking path for tokens generated under a superseded policy is implemented and unit-verified.

\cparagraph{Limitations} \CSB's state space assumes a verifiable, effectively binary reward, since the calibration buckets are indexed by the bit $c$; a reward model would need a scalar disagreement statistic instead, which we have not evaluated. Our workloads are math-only and colocated on H20 GPUs. Each configuration is a single run, so Table~\ref{tab:model-reward} carries no confidence interval, and a $20$-step horizon is short enough that slow divergence would not show in a final checkpoint, which is why \S\ref{sec:online-results} tracks \texttt{task\_score} step by step instead.

\cparagraph{Future work} Resumed groups supply $18.3$ percentage points of accepted groups on the main run, the gap between the $98.9\%$ and $80.6\%$ recalls of Table~\ref{tab:online-arc}. Whether a group whose samples straddle an update distorts its own group-relative advantage is not directly measured, and a fresh-reroll comparison on reward, KL, and entropy would settle this question most directly. Park-pool capacity is currently swept offline and static, though the right value depends on how much recoverable work the workload offers, and the controller could set it online from recent pool occupancy, applying the same closed-loop principle by which \AAP already sets the admission count. Separately, the joint-stop probability (Equation~\ref{eq:joint-probability}) treats candidate quality as independent; a correlated estimate would remove the reliance on conservative fallback under heavy simultaneous replacement demand.

\section{Related Work}
\label{sec:related-work}

\cparagraph{RL frameworks and cross-stage execution} veRL~\cite{sheng2025hybridflow} and slime~\cite{slime_github} integrate training and inference engines; later systems make placement dynamic, routing agentic stages to hardware-affine pools~\cite{gao2026rollart}, co-scheduling jobs to reclaim synchronization bubbles~\cite{wu2026weave,wu2025rollmux}, or migrating resources as bottlenecks shift~\cite{wang2026dynarl,yu2026rlinf}; a further group relaxes the barrier under bounded staleness~\cite{fu2026areal,zhong2025streamrl,han2025asyncflow,wang2025distflow}. All optimize across stages or jobs, whereas \SystemName takes placement and the barrier as given and asks which candidates within one rollout phase are worth computing at all.

\cparagraph{Rollout scheduling and tail mitigation} Seer approximates longest-job-first scheduling from an early probe response~\cite{qin2026seer}, and leverages divided rollout and adaptive grouped speculative decoding to accelerate decoding. RollPacker concentrates long-tail responses into designated long steps~\cite{gao2025rollpacker}, and TailSieve routes them to an isolated low-concurrency pool~\cite{xu2026tailsieve}. All run every admitted sample to completion, including the oversampled groups the filter will later discard, changing only how and where work executes, never how much. \S\ref{sec:priority-ceiling} bounds this class of intervention at $1.22\times$ even with perfect length knowledge.

\cparagraph{Early stopping and partial rollout}
Prior work uses partial rollout information to alter which trajectories are computed: SARA~\cite{voss2026early} commits effective groups early or abandons saturated ones, Selective Rollout~\cite{zhai2026selective} terminates groups predicted to have zero-variance rewards, and DARTS~\cite{wang2026darts} samples selectively from a redundant pool. Separately, slime~\cite{slime_github} and AReaL~\cite{fu2026areal} let an aborted generation resume later, and April~\cite{zhou2025april} preempts long-tail generations to keep the batch moving. \SystemName\ reuses this park and resume infrastructure but at group granularity and under a different invariant: it parks a whole group because a learned stop was uncertain, making that decision recoverable rather than final, resumes it only while it still looks likely to be kept, and requires all $S$ samples again before any reach the update, masking out those completed under an older policy version. No partially generated trajectory reaches policy optimization, so the group-shaped interface survives an asynchronous-style mechanism.
\section{Conclusion}
\label{sec:conclusion}

We present \SystemName, a closed-loop control architecture that reduces makespan and wasted generation in synchronous LLM RL rollout. Its core idea is to decide, from a group's partial completions, whether that group is still worth completing, and to make that decision recoverable rather than terminal. Experimental results show that \SystemName reduces rollout time by $37.4\%$, a $1.60\times$ speedup, and recovers $98.9\%$ of the groups that would ultimately have been kept, and preserves downstream accuracy within statistical noise. These results show that closing the loop around continuation, recovery, and admission is an effective mechanism for removing wasted generation from synchronous LLM RL rollout.


\bibliographystyle{ACM-Reference-Format}
\bibliography{references}

\end{document}